\documentclass[10pt,twocolumn,letterpaper]{article}

\usepackage[pagenumbers]{iccv} 
\usepackage{tabularx}

\definecolor{iccvblue}{rgb}{0.21,0.49,0.74}
\usepackage[pagebackref,breaklinks,colorlinks,allcolors=iccvblue]{hyperref}

\def\paperID{*****} 
\def\confName{ICCV}
\def\confYear{2025}

\title{Improving the physics of video generation with VJEPA-2 reward signal
}

\author{
Jianhao Yuan$^{1,2,}$\thanks{Work done during internship at Meta.} \, Xiaofeng Zhang$^{1,3,4}$ \, Felix Friedrich$^{1}$ \, Nicolas Beltran-Velez$^{1,5,\ast}$ \\ Melissa Hall$^{1}$ \, Reyhane Askari-Hemmat$^{1}$ \, Xiaochuang Han$^{1}$, Nicolas Ballas$^{1}$ \\ Michal Drozdzal$^{1,\dagger}$ \, Adriana Romero-Soriano$^{1,3,6,7,}$\thanks{Senior co-leads.}\\ \\
$^{1}$FAIR, Meta Superintelligence Labs \, $^{2}$University of Oxford\, $^{3}$Mila - Qu\'{e}bec AI Institute \\ $^{4}$Universit\'{e} de Montr\'{e}al \, $^{5}$Columbia University \, $^{6}$McGill University \, $^{7}$Canada CIFAR AI Chair\\
{\tt\small \{adrianars, mdrozdzal\}@meta.com, jianhaoyuan@robots.ox.ac.uk}}

\begin{document}
\maketitle

\emph{This is a short technical report describing the winning entry of the PhysicsIQ Challenge, presented at the Perception Test Workshop at ICCV 2025.}

\section{Introduction}
\label{sec:intro}
State-of-the-art video generative models~\citep{magi1, cogvideox, sora, videopoet, lumiere, wan2025} exhibit severely limited physical understanding, and often produce implausible videos. The Physics IQ benchmark~\citep{physicsIQ} has shown that visual realism does not imply physics understanding. Yet, intuitive physics understanding has shown to emerge from SSL pretraining on natural videos~\citep{vjepa2}. In this report, we investigate whether we can leverage SSL-based video world models to improve the physics plausibility of video generative models. 

In particular, we build ontop of the state-of-the-art video generative model MAGI-1~\citep{magi1} and couple it with the recently introduced Video Joint Embedding Predictive Architecture 2 (VJEPA-2) to guide the generation process~\citep{magi1}. We show that by leveraging VJEPA-2 as reward signal, we can improve the physics plausibility of state-of-the-art video generative models by $\sim6\%$.

\section{Background}
\label{sec:back}

\paragraph{MAGI-1~\cite{magi1}.} MAGI-1 is an autoregressive video generative model that predicts a sequence of video chunks. The model is trained to iteratively denoise each chunk and enables casual temporal video modeling. The model is able to generate videos when conditioned on text, text and image, as well as text and video.

At the inference time, the model uses the following score function to guide the diffusion process: 
\begin{align*}
\nabla_{x_t} \textrm{log} \ p_s (x_t | x_{<t}, \textrm{txt}) = (1- \omega_{<t}) \nabla_{x_t} \textrm{log} \ p(x_t) \\ + \  (\omega_{<t} - \omega_{\textrm{txt}}) \nabla_{x_t} \textrm{log} \ p(x_t | x_{<t}) \\ + \ \omega_{\textrm{txt}} \nabla_{x_t} \textrm{log} \ p(x_t | x_{<t}, \textrm{txt}),   
\label{eq:score}
\end{align*}
 where $x_t$ is the current video chunk, $x_{<t}$ are the past video chunks, txt is the text conditioning, $\omega_{<t}$ is the context guidance weight and $\omega_{\textrm{txt}}$ is the text guidance weight. 

\paragraph{VJEPA-2~\cite{vjepa2}.}  VJEPA-2 is a self-supervised video model which consists of a video encoder, $E_\theta$, and a predictor, $P_{\phi}$, trained with a Internet-scale data using a mask-denoising objective in the representation space.\looseness-1

\section{Method}
\label{sec:method}

We use VJEPA-2 as reward signal during the generation process of the state-of-the-art image-to-video (I2V) and video-to-video (V2V) MAGI-1 model. To do so, we leverage the \emph{surprise score} introduced in~\citep{vjepa2} and compare the generations of MAGI-1 with the VJEPA-2 predictions as follows:\looseness-1

\begin{equation}
S(x_t | x_{<t}, \textrm{txt}) = \textrm{sim}\big(P_{\phi}(E_\theta(x_{<t})), E_\theta(G_\gamma(x_{<t}, \textrm{txt})\big),
\label{eq:surprise}
\end{equation}

where $E_\theta$ and $P_{\phi}$ are the VJEPA-2 encoder and predictor, respectively; $G_{\gamma}$ is the generative model; and $\textrm{sim}$ is the cosine similarity.

We start by altering the score function of MAGI-1 to include signals from VJEPA-2 as follows:

\begin{equation}\label{eq:score}
    \begin{aligned}
\nabla_{x_t} \textrm{log} \ p_s (x_t | x_{<t}, \textrm{txt}) = (1- \omega_{<t}) \nabla_{x_t} \textrm{log} \ p(x_t) \\ + \  (\omega_{<t} - \omega_{\textrm{txt}}) \nabla_{x_t} \textrm{log} \ p(x_t | x_{<t}) \\ + \ \omega_{\textrm{txt}} \nabla_{x_t} \textrm{log} \ p(x_t | x_{<t}, \textrm{txt}) \\ - \ \omega_s \nabla_{x_t} S(x_t | x_{<t},\textrm{txt}) 
\end{aligned}
\end{equation}

Following Eq.~\eqref{eq:score}, we collect 16 video samples per condition. Then, we apply Best of N (BoN) and select the video sample with the lowest average surprise score Eq.~\eqref{eq:surprise}.\looseness-1

\section{Results}
\label{sec:results}

We evaluate our approach on the PhysicsIQ dataset and report the results in Table~\ref{tab:physicsiq}. By leveraging both VJEPA-2's surprise score as guidance and VJEPA-2 signal to select BoN, our method boosts vanilla sampling of MAGI-1 model and reaches a new state-of-the-art result of \textbf{62.64} and \textbf{36.86} on V2V and I2V, respectively.


\begin{table}[ht]
\centering
\caption{Evaluation on \textbf{PhysicsIQ}}
\begin{tabularx}{\linewidth}{lc}
\toprule
& PhysicsIQ Score ($\uparrow$)\\
\midrule
\multicolumn{2}{l}{\emph{V2V Generation}} \\
\midrule
Lumiere~\citep{lumiere}  & 23.00 \\
VideoPoet~\citep{videopoet} & 29.50 \\
MAGI-1~\citep{magi1}   & 56.31 \\
\midrule
MAGI-1 + VJEPA-2 (ours) & \textbf{62.64}	(+6.33) \\
\midrule
\multicolumn{2}{l}{\emph{I2V Generation}} \\
\midrule
Sora~\citep{sora}       & 10.00 \\
Pika 1.0~\citep{pika_1.0_2023}  &  13.00 \\
SVD~\citep{svd} & 14.80 \\
Lumiere~\citep{lumiere}   & 19.00 \\
VideoPoet~\citep{videopoet} & 20.30 \\
Wan2.1~\citep{wan2025}    & 20.89 \\
Runway Gen 3~\citep{runway_gen3_alpha_2024}     & 22.80 \\
CogVideoX~\citep{cogvideox} & 26.22 \\
MAGI-1~\citep{magi1}    & 30.23 \\
\midrule
MAGI-1 + VJEPA-2 (ours) & \textbf{36.86} (+6.63) \\
\bottomrule
\end{tabularx}
\label{tab:physicsiq}
\end{table}

{
    \small
    \bibliographystyle{ieeenat_fullname}
    \bibliography{main}
}

\end{document}